\documentclass[11pt,a4paper]{article}
\usepackage[hyperref]{acl2018}
\usepackage{times}
\usepackage{latexsym}

\usepackage{url}
\usepackage{graphicx}
\usepackage{enumitem}
\usepackage{multirow}
\usepackage{array}
\usepackage{todonotes}
\usepackage{subcaption}
\usepackage{hyperref}

\setlist[itemize]{leftmargin=*}

\aclfinalcopy % Uncomment this line for the final submission
\newcommand*\samethanks[1][\value{footnote}]{\footnotemark[#1]}

\newif\ifblindreview \blindreviewfalse

\title{The Best of Both Worlds: \\Combining Recent Advances in Neural Machine Translation\\
~}

\author{
Mia Xu Chen \thanks{~~~Equal contribution.}\\\And 
Orhan Firat \samethanks[1] \And 
Ankur Bapna \samethanks[1] \vspace*{-10cm} \\\AND 
~~~~~~~~~~~~Melvin Johnson\\ \And
~~~~~~~~~~~~~~~~~~~~Wolfgang Macherey\\\And 
~~~~~~~~~~~~~~~~~~~~~~~~George Foster\\\And 
~~~~~~~~~~~~~Llion Jones\\\And
 Niki Parmar \\\AND 
~~~~~~~~~~~~~~~~~~~~~~~~~~~~~~~Mike Schuster\\\And
~~~~~~~~~~~~~Zhifeng Chen\vspace{0.2cm}\\
~~~~~~~~~~~~~~~~~~~~~~~~~~~~~~~~~~~~~~~~~{\tt miachen,orhanf,ankurbpn,yonghui@google.com }\vspace{0.2cm}\\
~~~~~~~~~~~~~~~~~~~~~~~~~~~~~~~~~~~Google AI\\
~\\~
\And 
\hspace{-1.2cm} Yonghui Wu\\\And
\hspace{-3.2cm}Macduff Hughes
}

\begin{document}
\maketitle

\begin{abstract}

The past year has witnessed rapid advances in sequence-to-sequence (seq2seq)
modeling for Machine Translation (MT). The classic RNN-based approaches to MT
were first out-performed by the convolutional seq2seq model, which was then
out-performed by the more recent Transformer model. Each of these new
approaches consists of a fundamental architecture accompanied by a set of
modeling and training techniques that are in principle applicable to other
seq2seq architectures. In this paper, we tease apart the new architectures and
their accompanying techniques in two ways. First, we identify several key
modeling and training techniques, and apply them to the RNN architecture,
yielding a new RNMT+ model that outperforms all of the three fundamental architectures
on the benchmark WMT'14 English$\rightarrow$French and
English$\rightarrow$German tasks. Second, we analyze the properties of each
fundamental seq2seq architecture and devise new hybrid architectures intended
to combine their strengths. Our hybrid models obtain further improvements,
outperforming the RNMT+ model on both benchmark datasets.

\end{abstract}

\section{Introduction}

In recent years,  the emergence of seq2seq models \cite{kalchbrenner2013recurrent,sutskever2014sequence,ChoMGBSB14}
has revolutionized the field of MT by replacing traditional phrase-based approaches with
neural machine translation (NMT) systems based on the encoder-decoder paradigm. In the first
architectures that surpassed the quality of phrase-based MT, both
the encoder and decoder were implemented as Recurrent Neural Networks
(RNNs), interacting via a soft-attention mechanism \cite{BahdanauCB15}.
%, coupled with an attention mechanism.
%RNN based sequence-to-sequence models for Neural Machine Translation (RNMT)
%sparked off an upsurge in neural modelling, and machine translation
%has become a \textit{de facto} bechmarking problem ever since.
The RNN-based NMT approach, or RNMT, was quickly
established as the de-facto standard for NMT,
and gained rapid adoption into large-scale systems in industry,
e.g.~Baidu \cite{DBLP:journals/corr/ZhouCWLX16},
Google \cite{DBLP:journals/corr/WuSCLNMKCGMKSJL16},
and Systran \cite{DBLP:journals/corr/CregoKKRYSABCDE16}.
% [GF: alphabetizing for anonymity.]

%while attracting more
%attention toward neural modelling for MT. From the family of models perspective,
%RNMT was followed by convolutional based models. Fully convolutional
%sequence-to-sequence (ConvS2S) architectures were
%proposed in [cite], which outperformed the original RNMT based models in terms of
%translation quality while providing enhanced training speed due to the
%parallel computation within a sequence. [TODO(orhanf): we should
%  probably also talk a little bit about ByteNet here or before].

%Finding a better, more expressive, and more efficient architecture for
%sequence-to-sequence modeling has been at the heart of many
%research efforts, both in academia and industry.
Following RNMT, convolutional neural network based approaches
\cite{LeCun:1998:CNI:303568.303704} to NMT have recently drawn research
attention due to their ability to
fully parallelize training to take advantage of modern fast computing devices.
such as GPUs and Tensor Processing Units (TPUs) \cite{DBLP:journals/corr/JouppiYPPABBBBB17}.
%\todo{orhanf: citing TPU may hurt double blind, add this later?}
Well known examples are ByteNet
\cite{DBLP:journals/corr/KalchbrennerESO16} and ConvS2S
\cite{DBLP:journals/corr/GehringAGYD17}.
The ConvS2S model was shown to outperform the original RNMT architecture in terms
of quality, while also providing greater training speed.

Most recently, the Transformer model
\cite{DBLP:journals/corr/VaswaniSPUJGKP17}, which is based solely on
a self-attention mechanism \cite{Parikh2016ADA} and feed-forward connections, has further advanced
the field of NMT, both in terms of translation quality and speed of convergence.

% In many instances, however, improved
% performance often obfuscates the
% inescapable fact of neural modeling: that in order for them to work, each new model comes coupled with a `bag of
% tricks' which are only casually hinted at in
% publications or left for the keen researcher to discover in publicly released code (if any).
% These `tricks-of-the-trade' for neural networks hide an important
% scientific question, namely whether the observed performance
% improvements are primarely a result
% of the model structure itself or merely a consequence of the tricks
% that are invented or applied to make
% the model work or to further boost its performance. While the use of these tricks may sometimes be frowned upon
% by those who object the use of techniques that are neither fully
% understood nor explored with due diligence, employing these methods is often as
% crucial to the success of a newly advocated approach as the model itself.

%[GF: proposed replacement for the above. Is there a better term than `tricks'?]
In many instances, new architectures are accompanied by a novel set of
techniques for performing training and inference that have been carefully
optimized to work in concert. This `bag of tricks' can be crucial to the
performance of a proposed architecture, yet it is typically under-documented and
left for the enterprising researcher to discover in publicly released code (if
any) or through anecdotal evidence. This is not simply a problem for
reproducibility; it obscures the central scientific question of how much of the
observed gains come from the new architecture and how much can be
attributed to the associated training and inference techniques. In
some cases, these new techniques may be broadly applicable to other
architectures and thus constitute a major, though implicit, contribution
of an architecture paper. Clearly, they need to be considered in order
to ensure a fair comparison across different model architectures.

% [TODO(orhanf): we should probably emphasize these tricks are super important
% and equally valuable, since we are dealing with end-to-end nonlinear dynamical
% systems at the end.]
% [TODO(orhanf): i think i couldn't emphasize the applicability of tricks on
% different models, help!]
In this paper, we therefore take a step back and look at which
techniques and methods contribute
significantly to the success of recent architectures, namely ConvS2S and
Transformer, and explore applying
these methods to other architectures, including RNMT models.
%In doing so, we come up with the next generation RNN based NMT architecture,
%referred to as RNMT, that matches or exceeds the performance
%of current state-of-the-art approaches.
In doing so, we come up with an enhanced version of RNMT, referred to as RNMT+,
that significantly outperforms all individual architectures in our setup.
We further introduce new
architectures built with different components borrowed from
RNMT+, ConvS2S and Transformer.
%% We quantify the effect of each of these techniques with ablation experiments, both
%% in terms of the original model architecture and the new RNMT+ architecture.
%% We demonstrate that these techniques are equally applicable across different model
%% architectures.
%Inspired by recent developments, RNMT
%incorporates architectural ideas and training techniques from Transformer model.
%We demonstrate that applying these methods to RNMT leads to similar
%quality improvements and brings them en par with or even outperforms
%Transformers in terms of translation quality.
In order to ensure a fair setting for comparison, all architectures
were implemented in the same framework, use the same
pre-processed data and apply no further post-processing as this may
confound bare model performance.

%% Inspired by our understanding of the relative strengths and weaknesses
%% of individual model architectures, we further introduce a few new
%% architectures built with different components borrowed from
%% RNMT+, ConvS2S and Transformer.
%These hybrid models establish new
%state-of-the-art results on the standard WMT'14 English(En) $\rightarrow$ French(Fr)
%translation task and WMT'14 English $\rightarrow$ German(De) translation task.
% both of which are popular benchmark datasets in the MT research
% community.
Our contributions are three-fold:
\begin{enumerate}
\item In ablation studies, we quantify the effect of several modeling
improvements (including
multi-head attention and layer normalization) as well as
optimization techniques (such as synchronous replica training and
label-smoothing), which are used in recent architectures.
%, which are used in the recent Transformer model.
We demonstrate that these techniques are applicable
across different model architectures.

\item Combining these improvements with the RNMT model, we propose the new RNMT+
model, which significantly outperforms all fundamental architectures on
the widely-used WMT'14 En$\rightarrow$Fr
and En$\rightarrow$De benchmark datasets. We provide a detailed
model analysis and comparison of RNMT+, ConvS2S and Transformer
in terms of model quality, model size, and training and inference speed.

\item Inspired by our understanding of the relative strengths and weaknesses
of individual model architectures, we propose new model architectures that
combine components from the RNMT+ and the Transformer model, and achieve better
results than both individual architectures.
\end{enumerate}

We quickly note two prior works that provided empirical solutions
to the difficulty of training NMT architectures (specifically RNMT).
In \cite{britz-EtAl:2017:EMNLP2017}
the authors systematically explore which elements of NMT architectures have a
significant impact on translation quality.
%While both these works address important questions, we feel
%that another important question has been largely neglected.
In \cite{denkowski-neubig:2017:NMT}
the authors recommend three specific techniques for strengthening NMT systems
and empirically demonstrated how incorporating those techniques improves the
reliability of the experimental results.
% \todo{Summarize these findings, say how they relate to this paper?}

\section{Background}
\label{sec:background}

In this section, we briefly discuss the commmonly used NMT architectures.

%[TODO(orhanf): background is too verbose, formalize and simplify this entire
%section].

\subsection{RNN-based NMT Models - RNMT}
%% RNN based NMT models are the oldest and most commonly
%% used NMT architectures to date \cite{Forcada97}. In a nutshell, RNMT
%% architectures  are constructed as a composite of an encoder RNN and a decoder
%% RNN, coupled with an attention network to connect the decoder to the encoder.
RNMT models are composed of an encoder RNN and a decoder RNN, coupled with an
attention network. The encoder summarizes the input sequence into a set of
vectors while the decoder conditions on the encoded input sequence through an
attention mechanism, and generates the output sequence one token at a time.

%, and the
%decoder RNN uses the encoded set of vectors to generate the output
%sequence.

%In practice, the encoder is a recurrent language model over the input
%sequence \cite{mikolov2010recurrent} and the decoder is a recurrent conditional language model [cite Jamie]
%over the output sequence, where the conditioning is on the information coming
%from the encoder. There are two major branches in RNN based NMT models depending
%on how the decoder uses the encoded information. Decoders that utilize the
%encoded input information with a fixed vector [Ilya] -- usually the last hidden
%state of the encoder RNN, and decoders that are equipped with an additional
%attention network. This attention network attends over the encoded set of vectors
%and dynamically summarizes the encoded set into a context vector which the decoder RNN
%uses to generate the output sequence [Bahdanau]. The second branch of the RNN based NMT
%models, usually called as Attention based NMT, relaxes the information
%bottleneck between encoder and decoder, leading to better performance over
%the first branch.

%In this paper, we focus on the second branch of RNN based attentional NMT models
%and propose the next generation RNN based NMT
%models (RNMT+).

The most successful RNMT models consist of stacked RNN encoders with one or
more bidirectional RNNs, and stacked decoders with unidirectional RNNs. Both
encoder and decoder RNNs consist of either LSTM \cite{hochreiter1997long} or
GRU units \cite{ChoMGBSB14}, and make extensive use of residual
\cite{DBLP:journals/corr/HeZRS15} or highway
\cite{DBLP:journals/corr/SrivastavaGS15} connections.
%In order to
%narrow down the search space and build on top of a strong RNMT model,
%we picked the state-of-the-art RNMT model as the foundation of our RNMT+
%model, which is known as Google-NMT architecture [gnmt].

In Google-NMT (GNMT) \cite{DBLP:journals/corr/WuSCLNMKCGMKSJL16},
the best performing RNMT model on the datasets we consider,
the encoder network consists of one bi-directional LSTM layer, followed by 7
uni-directional LSTM layers. The decoder is equipped with a single attention
network and 8 uni-directional LSTM layers.
% The attention network is implemented as a single hidden layer feed forward
% neural network with $\tanh$ activations, and uses hidden states from the bottom
% decoder layer as the query to extract a context vector from the encoder final
% layer output.  The context vector is then used by each decoder RNN layer as an
% additional input to the hidden states from the previous RNN layer.
Both the encoder and the decoder use residual skip connections between
consecutive layers.

In this paper, we adopt GNMT as the starting point for our proposed RNMT+ architecture,
following the public NMT codebase\footnote{https://github.com/tensorflow/nmt}.

% [TODO(orhanf): explain this more].

%[TODO(orhanf): talk about RNN modeling the state space, good for sequence
%modeling especially for language]

%For problems where modeling the state information is essential, as in a language model, or in
%problems where modeling the phrase trajectory is beneficial,
%RNN based architectures excel, as they are dynamical systems. They can
%also be advantageous in auto-regressive models, owing to the infinite Markov structure of RNNs.

\subsection{Convolutional NMT Models - ConvS2S}

In the most successful convolutional sequence-to-sequence model
\cite{DBLP:journals/corr/GehringAGYD17}, both the encoder and decoder
are constructed by stacking multiple convolutional layers, where each
layer contains 1-dimensional convolutions followed by a gated linear units
(GLU) \cite{DBLP:journals/corr/DauphinFAG16}.
%The input to each layer
%is padded to ensure that the length of the layer output matches that
%of the layer
%input. This is done on both sides for the encoder layers but only on the left
%side for the decoder layers.
Each decoder layer computes a separate dot-product
attention by using the current decoder layer output and the final encoder layer
outputs. Positional embeddings are used to provide explicit positional information to the model.
%added to the token embeddings as network
%input, as well as to the final encoder layer output where they form
%the value vectors for the attention mechanism.
Following the practice in \cite{DBLP:journals/corr/GehringAGYD17}, we scale the
gradients of the encoder layers to stabilize training. We also use residual
connections across each convolutional layer and apply weight normalization
\cite{DBLP:journals/corr/SalimansK16} to speed up convergence. We follow
the public ConvS2S codebase\footnote{https://github.com/facebookresearch/fairseq-py} in our experiments.

%Linear projections are
%added to the embedding layer
%output, final encoder layer output, attention query vectors, attention context
%vectors, final decoder layer output before softmax, and between any two
%adjacent convolutional layers that have different hidden dimensions.

%[TODO(orhanf): talk about conv. modeling local context and correlations,
%lacking the state information].

% \enlargethispage*{.5cm}

\subsection{Conditional Transformation-based NMT Models - Transformer}
\label{subsec:transf}

The Transformer model \cite{DBLP:journals/corr/VaswaniSPUJGKP17} is motivated
by two major design choices that aim to address deficiencies in the former two
model families: (1) Unlike RNMT, but similar to the ConvS2S, the
Transformer model avoids any sequential dependencies in both the encoder and
decoder networks to maximally parallelize training. (2) To address the limited
context problem (limited receptive field) present in ConvS2S,
the Transformer model makes pervasive use of self-attention networks
\cite{Parikh2016ADA} so that
each position in the current layer has access to information from all other
positions in the previous layer.

%The latest family of NMT models try to overcome the deficits of its
%predecessors. As mentioned above, RNMT processes its input and output
%sequences with a recurrence relation, inherently depending on the
%representation at time step $i-1$, before processing time step
%$i$. This sequential dependency hurts training speed and prevents
%effective parallelization. On the other hand, by giving up on the
%infinite Markovian assumption of RNNs, convolutional models can enjoy
%massive parallelism, increasing the throughput of the network.  The
%disadvantage of convolutional models is their limited receptive field
%due to the very convolution operation, which can be mitigated by
%stacking more layers, and gradually increasing the receptive field of
%the entire encoder and decoders stacks at each added layer. However,
%as models get deeper and more expressive their trainability
%diminishes.

%Transformer \cite{DBLP:journals/corr/VaswaniSPUJGKP17} addresses the two above mentioned deficiencies,
%sequential processing and capturing only local context, by making use of
%pervasive attention networks rather than convolutions and avoiding any
%recurrence relations. [TODO(orhanf): trade horizontal dependency with vertical
%dependency, which is practical when depth ellell seqlen]

The Transformer model still follows the encoder-decoder paradigm. Encoder
transformer layers are built with two sub-modules: (1) a self-attention network
and (2) a feed-forward network. Decoder transformer layers have an additional
cross-attention layer sandwiched between the self-attention and feed-forward
layers to attend to the encoder outputs.
%The self-attention network replaces the affine transformation between consecutive
%layers in conventional feed-forward networks with an attention network, where
%each layer activation becomes a transformed weighted average of all the
%activations of the previous layer.
% The clever use of self-attention solves two problems: (a) it
% avoids any recurrent connections as in RNMT models and (b) it mitigates the local
% context problem as present in the ConvS2S model.
%Compared to the attention networks that
%are used between encoder and decoder in RNMT or ConvS2S, the self-attention network
%does not rely on a query from the decoder or any other external layer,
%but rather the layer that it is attending, lending the name self-attention.
%This allows the forward propagating network to have access to any activation
%of the layer below, mitigating the local context problem.
%By making
%clever use of activations of the layer being processed as queries, self-attention
%also refrains from using any recurrence relations over the input sequence,
%but only between layers.
% The self-attention module is followed by a
% feed-forward network to increase the modeling capacity of the network.
%The decoder stack in a Transformer network follows the structure of the encoder
%and is again composed by 6 transformer layers.
% On the decoder side, each layer has an extra cross-attention layer that is
% placed between the self-attention layer and the feed-forward layer and
% attends  to the encoder outputs.

%A couple of very important but often ignored components need further
%elaboration in the Transformer.

There are two details which we found very important to the model's
performance:  (1) Each sub-layer in the transformer (i.e.~self-attention,
cross-attention, and the feed-forward sub-layer) follows a strict computation
sequence: \textit{normalize} $\rightarrow$ \textit{transform} $\rightarrow$
\textit{dropout}$\rightarrow$ \textit{residual-add}.
(2) In addition to per-layer normalization, the final encoder output is again
normalized to prevent a blow up after consecutive residual additions.
%These
%last two wiring changes on top of the original Transformer network qualify as
%an enhanced version.

In this paper, we follow the latest version of the Transformer model in the
public Tensor2Tensor\footnote{https://github.com/tensorflow/tensor2tensor} codebase.

%In addition to the
%normalization first processing sequence, after stacking 6 transformer layers
%in the encoder, the entire encoder block output is again normalized, in order
%not to blow up after consecutive residual additions.\footnote{Note that, both
%above enhancements are scraped from the tensor2tensor public codebase}
%Conditions (loosely speaking) on an additinoal query, as it transforms it's input.

\begin{figure*}[t!] \centering
\includegraphics[width=0.8\textwidth,height=7cm]{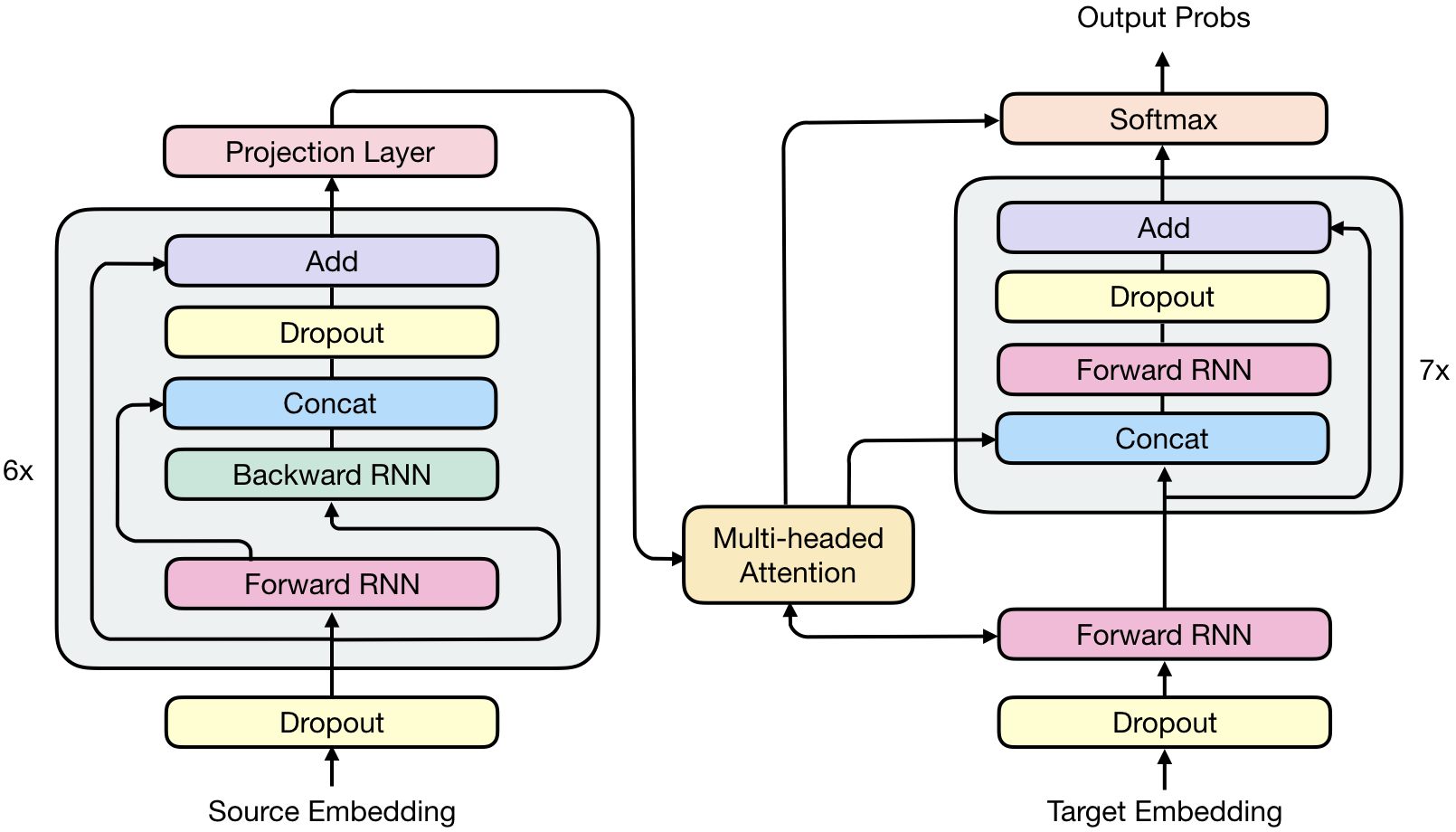} \caption{Model
architecture of RNMT+. On the left side, the encoder network has 6 bidirectional
LSTM layers. At the end of each bidirectional layer, the outputs of the forward
layer and the backward layer are concatenated.  On the right side, the decoder
network has 8 unidirectional LSTM layers, with the first layer used for
obtaining the attention context vector through multi-head additive attention. The
attention context vector is then fed directly into the rest of the decoder
layers as well as the softmax layer.} \label{fig:gnmtv2} \end{figure*}

% \subsection{A Theory-based analysis of NMT Architectures}
\subsection{A Theory-Based Characterization of NMT Architectures}

From a theoretical point of view, RNNs belong to the most expressive
members of the neural network family \cite{SIEGELMANN1995132}\footnote{Assuming that data
complexity is satisfied.}. Possessing an infinite Markovian structure (and thus
an infinite receptive fields) equips them to model sequential data \cite{ELMAN1990179},
especially natural language \cite{Grefenstette:2015:LTU:2969442.2969444} effectively.
In practice, RNNs are notoriously hard to train \cite{Bengio:1994:LLD:2325857.2328340},
confirming the well known dilemma of trainability versus expressivity.
%Recent
%work states that, when trained effectively \cite{Collins2016CapacityAT},
%the theoretical power of RNNs can be revealed \cite{DBLP:journals/corr/MelisDB17},
%but unfortunately this has not yet been validated for MT.

Convolutional layers are adept at capturing local context and local
correlations by design. A fixed and narrow receptive field for each
convolutional layer limits their capacity when the architecture is shallow. In
practice, this weakness is mitigated by stacking
more convolutional layers (e.g.~15 layers as in the
ConvS2S model), which makes the model harder to train and
demands meticulous initialization schemes and carefully designed
regularization techniques.

%The ConvS2S model ameliorates this weakness
% of having narrow-receptive field,
%by stacking more convolutional layers, which expands their receptive
%fields with each added convolutional layer. As mentioned in the
%previous section, as the models get deeper and more expressive, ease
%of trainability diminishes. We see that the ConvS2S architecture
%stacks 15 convolutional layers to expand the receptive fields in both
%encoder and decoder, which in turn makes the model harder to train and
%justifies the extensive use of weight normalization, explicit gradient
%scaling for the encoder and a very meticulous and specifically
%tailored initialization scheme tied with both the dropout scheme and
%the optimizer being used.

%The Transformer model is designed to bridge the gap between RNN and
%CNN families, expanding the receptive field to the sequence length by
%making use of attention, and can safely be put into the family of
%feed-forward neural networks with full connectivity.  Similar to our
%previous analysis, from a theoretical point\footnote{Given enough model
%  capacity and assuming that data complexity is satisfied}, the
The transformer network is capable of approximating arbitrary squashing
functions \cite{HORNIK1989359}, and can be considered a strong
feature extractor with extended receptive fields capable of linking salient
features from the entire sequence. On the other hand, lacking a memory component
(as present in the RNN models) prevents the network from modeling a state space,
reducing its theoretical strength as a sequence model, thus it requires
additional positional information (e.g. sinusoidal positional encodings).

Above theoretical characterizations will drive our explorations in the
following sections.

%This comparison with Turing Complete
%models (RNN) has to our knowledge not yet been done in a fair setup
%and on a large-scale, challenging task like MT.

% \todo{Need a few sentences to tie the story withe hybrid experiments, to
% motivate the choice of hybrid models we experimented with.}

\section{Experiment Setup}
\label{sec:eval}
\vspace{-5px}
We train our models on the standard WMT'14 En$\rightarrow$Fr and En$\rightarrow$De
datasets that comprise 36.3M and 4.5M sentence pairs, respectively.
Each sentence was encoded into a sequence of sub-word units obtained
by first tokenizing the sentence with the Moses tokenizer, then splitting tokens into sub-word
units (also known as ``wordpieces'')  using the approach described in \cite{wordpiece_schuster}.
%\cite{SennrichHB15}.

We use a shared vocabulary of 32K sub-word units for each
source-target language pair. No further manual or rule-based post
processing of the output was performed beyond combining the sub-word
units to generate the targets.  We report all our results on newstest
2014, which serves as the test set. A combination of newstest 2012 and
newstest 2013 is used for validation.

To evaluate the models, we compute the BLEU metric on tokenized, true-case
output.\footnote{This procedure is used in the literature to which we compare
  \cite{DBLP:journals/corr/GehringAGYD17,DBLP:journals/corr/WuSCLNMKCGMKSJL16}.}
For each training run, we evaluate the
model every 30 minutes on the dev set. Once the model converges, we
determine the best window based on the average dev-set BLEU score over
21 consecutive evaluations.
% [GF: replicability problem: people would need to know how fast you trained in
% order to match your window size. Should give a rough equivalent in # examples.]
We report the mean test score and standard
deviation over the selected window. This allows us to compare model
architectures based on their mean performance after convergence rather than
individual checkpoint evaluations, as the latter can be quite noisy for some models.

To enable a fair comparison of architectures, we use the same pre-processing and
evaluation methodology for all our experiments. We refrain from using checkpoint
averaging (exponential moving averages of parameters) \cite{junczys2016amu} or
checkpoint ensembles \cite{jean2015using,DBLP:journals/corr/abs-1710-03282}
to focus on evaluating the performance of individual models.

% As some models in the literature
% report final BLEU scores after checkpoint averaging, and decoding with the
% averaged checkpoint, we refrain from using such techniques because chekpoint
% averaging (exponential moving averages of parameters)\cite{junczys2016amu},
% or checkpoint ensembles \cite{jean2015using,DBLP:journals/corr/abs-1710-03282}
% can all be considered as post-processing steps and may hide the bare
% performance of the model at hand.

\section{RNMT+}
\label{sec:v2}

\subsection{Model Architecture of RNMT+}
The newly proposed RNMT+ model architecture is shown in Figure~\ref{fig:gnmtv2}.
Here we highlight the key architectural choices that are different between
the RNMT+ model and the GNMT model.
There are 6 bidirectional LSTM layers in
the encoder instead of 1 bidirectional LSTM layer followed by 7 unidirectional
layers as in GNMT. For each
bidirectional layer, the outputs of the forward layer and the backward layer
are concatenated before being fed into the next layer.
The decoder network consists of 8 unidirectional LSTM layers similar to the
GNMT model.
Residual connections are
added to the third layer and above for both the encoder and decoder. Inspired by
the Transformer model, per-gate
layer normalization \cite{DBLP:journals/corr/BaKH16} is applied within each LSTM cell. Our empirical results
show that layer normalization greatly stabilizes training. No non-linearity is
applied to the LSTM output.  A projection layer is added to the encoder final
output.\footnote{Additional projection aims to reduce the dimensionality of
the encoder output representations to match the decoder stack dimension.}
Multi-head additive attention is used instead of the single-head
attention in the GNMT model. Similar to GNMT, we use the bottom decoder layer
and the final encoder layer output after projection for obtaining the recurrent
attention context. In addition to feeding the attention context to all decoder
LSTM layers, we also feed it to the softmax.
This is important for both the
quality of the models with multi-head attention and the stability of the
training process.

Since the encoder network in RNMT+ consists solely of bi-directional LSTM
layers, model parallelism is not used during training. We compensate for
the resulting longer per-step time with increased data parallelism
(more model replicas), so that the overall time to reach
convergence of the RNMT+ model is still comparable to that of GNMT.

We apply the following regularization techniques during training.
\begin{itemize}
\item \textbf{Dropout:} We apply dropout to both embedding layers and each LSTM
 layer output before it is added to the next layer's input. Attention dropout is also applied.

\item \textbf{Label Smoothing:} We use uniform label smoothing with an
  uncertainty=0.1 \cite{DBLP:journals/corr/SzegedyVISW15}. Label smoothing was shown to have a positive impact on both
Transformer and RNMT+ models, especially in the case of RNMT+ with
multi-head attention. Similar to the observations in \cite{DBLP:journals/corr/ChorowskiJ16}, we found it beneficial to use a larger beam size
(e.g. 16, 20, etc.) during decoding when models are trained with label
smoothing.

\item \textbf{Weight Decay:} For the WMT'14 En$\rightarrow$De task, we apply L2
regularization to the weights with $\lambda = 10^{-5}$. Weight decay is only
applied to the En$\rightarrow$De task as the corpus is smaller and thus more
regularization is required.
\end{itemize}

We use the Adam optimizer \cite{DBLP:journals/corr/KingmaB14} with $\beta_1 = 0.9, \beta_2 = 0.999,
\epsilon = 10^{-6}$ and vary the learning rate according to this schedule:
\begin{equation}
lr = 10^{-4} \cdot \min\Big(1 + \frac{t \cdot (n-1)}{np}, n, n \cdot (2n)^{\frac{s- nt}{e - s}}\Big)
\label{eq:rnmt_lr}
\end{equation}
Here, $t$ is the current step, $n$ is the number of concurrent model replicas used in training, $p$
is the number of warmup steps, $s$ is the start step of the exponential decay, and
$e$ is the end step of the decay. Specifically, we first increase the learning
rate linearly during the  number of warmup steps, keep it a constant until the
decay start step $s$, then exponentially decay until the decay end step $e$,
and keep it at $5 \cdot 10^{-5}$ after the decay ends. This learning rate
schedule is motivated by a similar schedule that was successfully
applied in training the Resnet-50 model with a very large batch size
\cite{DBLP:journals/corr/GoyalDGNWKTJH17}.

In contrast to the asynchronous training used for GNMT
% in GNMT where models were trained with asynchronous training
\cite{downpoursgd}, we train RNMT+ models with synchronous training
\cite{DBLP:journals/corr/ChenMBJ16}. Our empirical results suggest that when
hyper-parameters are tuned properly, synchronous training often leads
to improved convergence speed and superior model quality.

To further stabilize training, we also use adaptive
gradient clipping. We discard a training step completely if an anomaly
in the gradient norm value is
detected, which is usually an indication of an imminent gradient explosion.
More specifically, we keep track of a moving average and a moving
standard deviation of the $\log$ of the gradient norm values, and we
abort a step if the norm of the gradient exceeds four standard deviations
of the moving average.

\subsection{Model Analysis and Comparison}

In this section, we compare the results of RNMT+ with ConvS2S and Transformer.

All models were trained with synchronous training.
RNMT+ and ConvS2S were trained with 32
NVIDIA P100 GPUs while the Transformer Base and Big models were trained using 16
GPUs.
% [GF: need to explain the diff between Base and Big.]

For RNMT+, we use sentence-level cross-entropy loss. Each training batch
contained 4096 sentence pairs (4096 source sequences and 4096 target sequences).
For ConvS2S and Transformer models, we use token-level cross-entropy loss. Each
training batch contained 65536 source tokens and 65536 target tokens. For the
GNMT baselines on both tasks, we cite the largest BLEU score reported in
\cite{DBLP:journals/corr/WuSCLNMKCGMKSJL16} without reinforcement learning.

Table~\ref{table:enfr} shows our results on the WMT'14 En$\rightarrow$Fr
task. Both the Transformer Big model and RNMT+ outperform GNMT and ConvS2S by
about 2 BLEU points. RNMT+ is slightly better than the Transformer Big model in
terms of its mean BLEU score. RNMT+ also yields a much lower standard
deviation, and hence we observed much less fluctuation in the training
curve. It takes approximately 3 days for the Transformer Base model to
converge, while both RNMT+ and the Transformer Big model require about
5 days to converge. Although the batching schemes are quite different
between the Transformer Big and the RNMT+ model, they have processed about the
same amount of training samples upon convergence.

\begin{table}[!htbp]
\centering
\begin{tabular}{ c|c|>{\centering\arraybackslash}m{1cm}|>{\centering\arraybackslash}m{1.1cm}}
 \hline
 \hline
Model & Test BLEU & Epochs & Training Time \\
 \hline
 GNMT & 38.95 & - & -\\
 ConvS2S \footnotemark[7] & 39.49 $\pm$ 0.11 & 62.2 & 438h\\
 Trans. Base & 39.43 $\pm$ 0.17& 20.7 & 90h\\
 Trans. Big \footnotemark[8] & 40.73 $\pm$ 0.19 & 8.3 & 120h\\
 RNMT+ & 41.00 $\pm$ 0.05 & 8.5 & 120h\\
 \hline
\end{tabular}
\caption{Results on WMT14 En$\rightarrow$Fr. The numbers before and after
'$\pm$' are the mean and standard deviation of test BLEU score over an
evaluation window.}
\label{table:enfr}
\end{table}

Table~\ref{table:ende} shows our results on the WMT'14 En$\rightarrow$De task.
The Transformer Base model improves over GNMT and ConvS2S by more than 2 BLEU
points while the Big model improves by over 3 BLEU points. RNMT+ further
outperforms the Transformer Big model and establishes a new state of
the art with an averaged value of 28.49. In this case, RNMT+ converged
slightly faster than the Transformer Big model and maintained much more
stable performance after convergence with a very small standard
deviation, which is similar to what we observed on the En-Fr task.

\footnotetext[7]{Since the ConvS2S model convergence is very slow we did
not explore further tuning on En$\rightarrow$Fr, and
validated our implementation on En$\rightarrow$De.}
\footnotetext[8]{The BLEU scores for Transformer model are slightly lower than those
reported in \cite{DBLP:journals/corr/VaswaniSPUJGKP17} due to four differences:

1) We report the mean test BLEU score using the strategy described in section~\ref{sec:eval}.

2) We did not perform checkpoint averaging since it would be inconsistent with our evaluation for other models.

3) We avoided any manual post-processing, like unicode normalization using Moses replace-unicode-punctuation.perl or output tokenization using Moses tokenizer.perl, to rule out its effect on the evaluation.
We observed a significant BLEU increase (about 0.6) on applying these post processing techniques.

4) In \cite{DBLP:journals/corr/VaswaniSPUJGKP17}, reported
BLEU scores are calculated using mteval-v13a.pl from Moses,
which re-tokenizes its input.}

\begin{table}[!htbp]
\centering
\begin{tabular}{ c|c|>{\centering\arraybackslash}m{1cm}|>{\centering\arraybackslash}m{1.1cm}}
 \hline
 \hline
Model & Test BLEU & Epochs & Training Time \\
 \hline
 GNMT & 24.67 & - & -\\
 ConvS2S &  25.01 $\pm$0.17 & 38 & 20h\\
 Trans. Base & 27.26 $\pm$ 0.15 & 38 & 17h\\
 Trans. Big & 27.94 $\pm$ 0.18 & 26.9 & 48h\\
 RNMT+ &  28.49 $\pm$ 0.05 & 24.6 & 40h\\
 \hline
\end{tabular}
\caption{Results on WMT14 En$\rightarrow$De.}
\label{table:ende}
\end{table}

Table~\ref{table:perf} summarizes training performance and model statistics.
The Transformer Base model is the fastest model in terms of training speed. RNMT+ is slower
to train than the Transformer Big model on a per-GPU basis.
However, since the RNMT+ model is quite stable, we were able to
offset the lower per-GPU throughput with higher concurrency by
increasing the number of model replicas, and
hence the overall time to convergence was not slowed down much.
We also computed the number of floating point operations (FLOPs)
in the model's forward path as well as the number of total parameters
for all architectures (cf.~Table~\ref{table:perf}). RNMT+ requires
fewer FLOPs than the Transformer Big model, even though both models
have a comparable number of parameters.

\begin{table}[!htbp]
\centering
\begin{tabular}{ c|>{\centering\arraybackslash}m{1.7cm}|>{\centering\arraybackslash}m{1.2cm}|>{\centering\arraybackslash}m{1.1cm}}
 \hline
 \hline
 Model & Examples/s & FLOPs & Params\\
 \hline
 ConvS2S & 80 & 15.7B &  263.4M\\
 Trans. Base & 160 & 6.2B &  93.3M\\
 Trans. Big &  50 & 31.2B & 375.4M\\
 RNMT+ & 30 & 28.1B & 378.9M\\
 \hline
\end{tabular}
\caption{Performance comparison. Examples/s are normalized by the number of
GPUs used in the training job. FLOPs are computed assuming that source
and target sequence length are both 50.}
\label{table:perf}
\end{table}

% \enlargethispage*{.5cm}
\vspace{-10px}
\section{Ablation Experiments}

In this section, we evaluate the importance of four main techniques
for both the RNMT+ and the Transformer Big models. We believe that these
techniques are universally applicable across different model
architectures, and should always be employed by NMT practitioners for
best performance.

We take our best RNMT+ and
Transformer Big models and remove each one of these techniques
independently. By doing this we hope to learn two things about each
technique: (1) How much does it affect the model performance? (2) How useful is
it for stable training of other techniques and hence the final model?

\begin{table}[!htbp]
\centering
\begin{tabular}{c|c|c}
 \hline
 \hline
 Model & RNMT+ & Trans. Big \\
 \hline
Baseline & 41.00 & 40.73\\
- Label Smoothing & 40.33 & 40.49\\
- Multi-head Attention & 40.44 & 39.83\\
- Layer Norm. & * & *\\
- Sync. Training & 39.68 & *\\
 \hline
\end{tabular}
\caption{Ablation results of RNMT+ and the Transformer Big model on WMT'14
En $\rightarrow$ Fr. We report average BLEU
scores on the test set. An asterisk '\mbox{*}' indicates an unstable training run
(training halts due to non-finite elements).}
\label{table:enfr_ablation}
\end{table}

From Table~\ref{table:enfr_ablation} we draw the following conclusions about the
four techniques:
\begin{itemize}
\item \textbf{Label Smoothing}
We observed that label smoothing improves both models, leading to an
average increase of 0.7 BLEU for RNMT+ and 0.2 BLEU for Transformer Big models.
\item \textbf{Multi-head Attention}
Multi-head attention contributes significantly to the quality of
both models, resulting in an average increase of 0.6 BLEU for RNMT+ and 0.9 BLEU
for Transformer Big models.
\item \textbf{Layer Normalization}
Layer normalization is most critical to stabilize the
training process of either model, especially when multi-head attention is used.
Removing layer normalization results in unstable training runs for both models.
Since by design, we remove one technique at a time in our ablation experiments,
we were unable to quantify how much layer
normalization helped in either case. To be able to successfully train a model
without layer normalization, we would have to adjust other parts of the model
and retune its hyper-parameters.
\item \textbf{Synchronous training}
Removing synchronous training has different effects on RNMT+ and Transformer.
For RNMT+, it results in a significant quality drop, while for the
Transformer Big model, it causes the model to become unstable. We also
notice that synchronous training is only successful when coupled
with a tailored learning rate schedule that has a warmup stage at
the beginning (cf.~Eq.~\ref{eq:rnmt_lr} for RNMT+ and
Eq.~\ref{eq:trans_lr} for Transformer). For RNMT+, removing this
warmup stage during synchronous training causes the model to become
unstable.
\end{itemize}

%\begin{table}[!htbp]
%\centering
%\begin{tabular}{ c|c}
% \hline
% \hline
%Beam Size &  Avg. Test BLEU\\
% \hline
% 8 & 40.91 \\
%12 & 41.00 \\
%16 & 40.96 \\
%20 & 40.90 \\
%24 & 40.93 \\
%30 & 40.92 \\
%40 & 40.79 \\
%50 & 40.58 \\
% \hline
%\end{tabular}
%\caption{The effect of beam size on the results of RNMT+ on WMT14 En $\rightarrow$ Fr. Similar delta in BLEU scores were observed on the dev
%set as well. [TODO(melvinp): There is enough space to report dev set
%  BLEU scores as well]}
%\label{table:beamsize}
%\end{table}
%
%Table~\ref{table:beamsize} shows how an increased beam size during
%inference affects the decoding quality of RNMT+. Similar to what was observed by \cite{DBLP:journals/corr/ChorowskiJ16} when label smoothing is used, a relatively larger beam size is
%beneficial. In our case, we reach the best translation quality using a beam size
%of between 12 and 20.
%
%[TODO(orhanf): maybe we can add a sequence length / performance table]

\section{Hybrid NMT Models}
\label{sec:hybrids}
\vspace{-5px}
In this section, we explore hybrid architectures that shed
some light on the salient behavior of each model family. These hybrid models
outperform the individual architectures on both benchmark datasets and provide
a better understanding of the capabilities and limitations of each model family.
\subsection{Assessing Individual Encoders and Decoders}
% \todo{better names for the two subsections?}

%In Sec.~\ref{sec:background}, we mentioned the expected salient behavior of a
%model that keeps track of a state information (RNN), and expected salient
%behavior of a model with expanded receptive field and easy access patterns
%for feature extraction (Feed-forward Networks with Self-Attention).

In an encoder-decoder architecture, a natural assumption
is that the role of an encoder is to build feature representations that can
best encode the meaning of the source sequence, while a decoder should
be able to process and interpret the representations from the encoder and,
at the same time, track the current target history.
% decode by the regularities of the target sequence.
% In the most powerful models,
Decoding is inherently auto-regressive,
and keeping track of the state information should therefore
%(following a trajectory
%in a semantic space or manifold)
be intuitively beneficial for conditional generation.
%On the other hand, coupling the encoder with the decoder via attention
%allows the decoder to access any position of the encoding and pick the most relevant information.
%This relaxes the
%necessity of forming distinct state information in the encoding,
%Allowing the attention network to pick most relevant representation from
%the encoder,
%but also requires the encoder to build rich and expressive representations.

%Following the above hypothesis,
%the very first question we ask as we start
%combining different components from each model is,
We set out to study which family of encoders
is more suitable to extract rich representations from a given input sequence,
and which family of decoders can make the best of such rich representations.
We start by combining the encoder and decoder from different
model families. Since it takes a significant amount of time for a
ConvS2S model to converge, and because the final translation quality
was not on par with the other models, we focus on two types of
hybrids only: Transformer encoder with RNMT+ decoder and RNMT+ encoder with
Transformer decoder.

\begin{table}[!htbp]
\centering
\begin{tabular}{c|c|c}
\hline \hline
Encoder & Decoder & En$\rightarrow$Fr Test BLEU \\ \hline
Trans. Big    & Trans. Big                    & 40.73 $\pm$ 0.19     \\
RNMT+    & RNMT+                    & 41.00 $\pm$ 0.05     \\
Trans. Big    & RNMT+                    & \textbf{41.12 $\pm$ 0.16}     \\
RNMT+     & Trans. Big          & 39.92 $\pm$ 0.21      \\ \hline
\end{tabular}
\caption{Results for encoder-decoder hybrids.}
\label{table:hybrids_encdec}
\end{table}

From Table~\ref{table:hybrids_encdec}, it is clear that the Transformer
encoder is better at encoding or feature extraction than the RNMT+
encoder, whereas RNMT+ is better at decoding or conditional language
modeling, confirming our intuition that a stateful decoder is
beneficial for conditional language generation.

% \paragraph{Stateful Decoding with Stateful Rich Features}
\subsection{Assessing Encoder Combinations}
%Given an empirical answer to our first question, a logical follow up question is,
Next, we explore how the features extracted by an encoder can be
further enhanced by incorporating additional
information. Specifically, we investigate the combination of
transformer layers with RNMT+ layers in the same encoder block to build even richer feature representations.
%richer in its abstraction and also keeping track of a state information at the same
%time. The question we want to answer is, can a stateful decoder benefit from
%additional encoder state information on top of rich features extracted from
%an input sequence.
We exclusively use RNMT+ decoders in the following architectures since stateful
decoders show better performance according to Table~\ref{table:hybrids_encdec}.
%Given that a rich feature extractor (or stacked geometric morphing) in the decoder
%degrades performance significantly (see second row of Table~\ref{table:hybrids_encdec}),
%we only focus on stateful decoders, namely RNMT+ as the decoder.

%Note that, the models in this section have increased model capacity due to increased
%\todo{orhanf: this paragraph needs a fix}
%number of parameters. But in our experiments with pure RNMT+ model and pure Transformer Big
%model, increasing the model capacity (increased number of encoder layers for RNMT+,
%increased number of encoder and decoder layers for Transformer Big), yielded worse
%performance. So our conclusions here in this section are not merely increased performance
%by increasing the model capacity.

%Given an empirical
%answer to our first question, a logical follow up question is, how can we improve
%the learned representations in a finer grained blending between components from
%different model families. To be specific, we next explore how can we combine/mix
%a transformer layer with an RNMT+ layer to have a computational block that is
%richer in its abstraction.

%Note that, in order to have valid experimental setup, we fixed everything but
%the tested hypothesis, therefore we stick to the RNMT+ decoder and only
%change the encoder block. This allows us to directly observe the effect of the
%above question asked.

\begin{figure}
\begin{subfigure}{.25\textwidth}
  \centering
  \includegraphics[width=.6\linewidth]{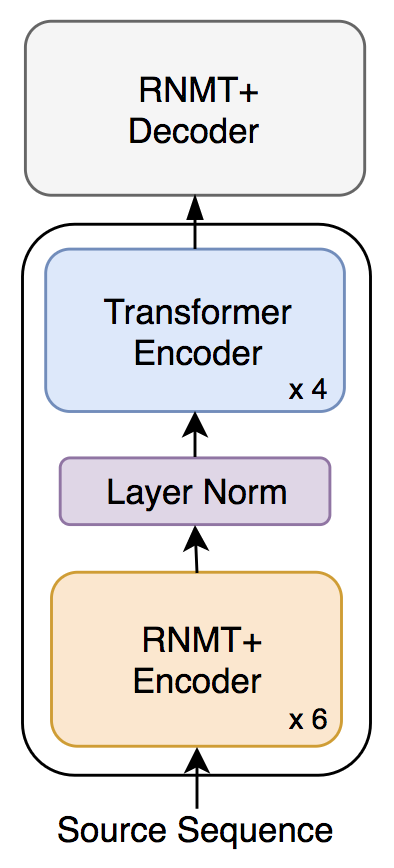}
  \caption{Cascaded Encoder}
  \label{fig:stacked}
\end{subfigure}%
\begin{subfigure}{.25\textwidth}
  \centering
  \includegraphics[width=.85\linewidth]{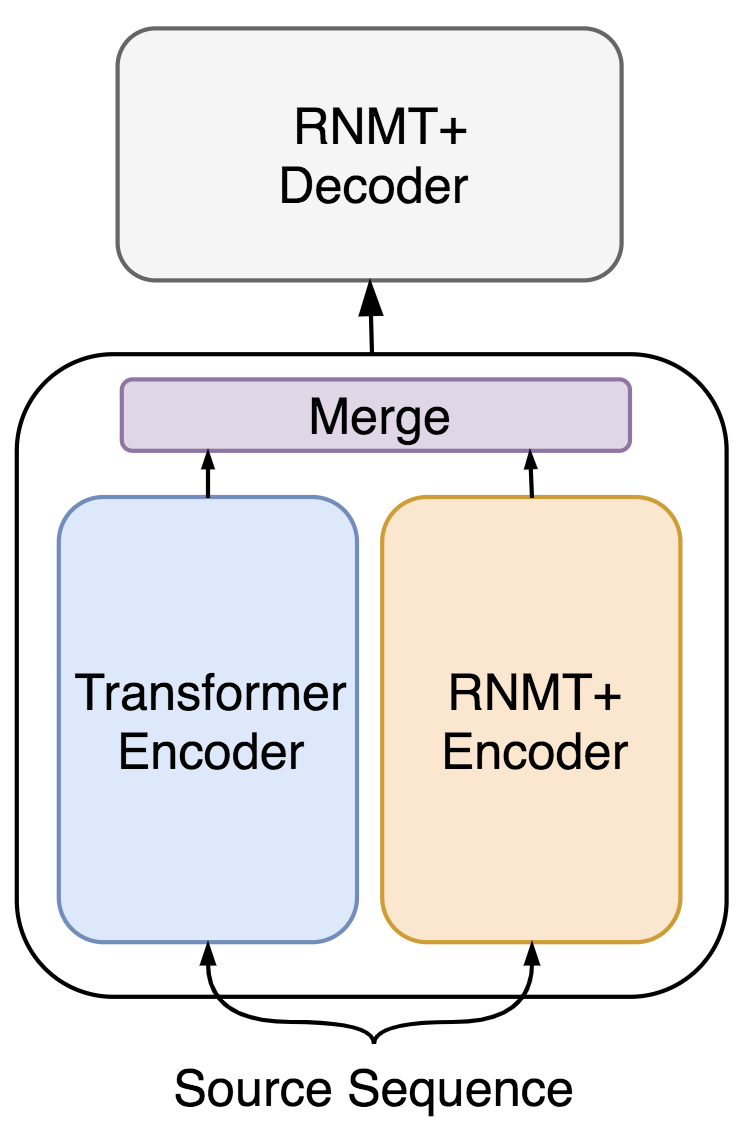}
  \caption{Multi-Column Encoder}
  \label{fig:mcol}
\end{subfigure}
\caption{Vertical and horizontal mixing of Transformer and RNMT+ components in an encoder.}
\label{fig:enc_hybrids}
\end{figure}

We study two mixing schemes in the encoder (see Fig.~\ref{fig:enc_hybrids}):

(1) \textit{Cascaded Encoder}: The cascaded encoder aims at combining
the representational power of RNNs and self-attention. The idea is to
enrich a set of stateful representations by cascading a feature
extractor with a focus on vertical mapping, similar to
\cite{DBLP:journals/corr/PascanuGCB13,D17-1300}.  Our best performing
cascaded encoder involves fine tuning transformer layers stacked on
top of a pre-trained frozen RNMT+ encoder.  Using a pre-trained
encoder avoids optimization difficulties while significantly enhancing encoder
capacity. As shown in Table~\ref{table:hybrids-perf},
the cascaded encoder improves over the Transformer encoder by more
than 0.5 BLEU points on the WMT'14 En$\rightarrow$Fr task. This
suggests that the Transformer encoder is able to extract richer
representations if the input is augmented with sequential context.
%with the hope that the two computational blocks compensate for the deficits of the other.
%In other words, can we enrich/boost a set of stateful representations,
%by further applying (cascading) a feature extractor that focuses on vertical
%mapping.

%As shown in Table~\ref{table:hybrids-perf}, RNMT+ decoder enjoys stateful and rich
%features from the encoder, compared to only stateful or only rich features alone.
%Note that, the ordering in the cascaded encoder matters, as in our experiments
%we observed degraded performance when we change the ordering (eg. adding state
%\todo{orhanf: ankur this needs your attention}
%information into rich features).

%encountered with non-frozen versions of the hybrid, and enhance encoder
%capacity significantly, resulting in improved performance as described in Table~\ref{table:hybrids-perf}.
%Our results suggest that the Transformer encoder, as a feature extractor,
%extracts richer representations, when the input sequence is augmented with sequential context.

%(1) \textit{Cascaded Encoder}. Vertical mixing (or cascading) is motivated by
%combining representatinal power of RNNs and self-attention by interleaving
%with the hope of two computational blocks meet the deficits of each other.
%and we stack a transformer
%encoder on top of the RNMT+ encoder. Although we experimented with several
%approaches to combine transformer, RNN and convolutional layers, our best
%performing vertical mixing hybrids are based on combinations of Transformer
%and BiLSRM layers, as shown in Table~\ref{table:hybrids-perf}.

(2) \textit{Multi-Column Encoder}:
%Next question we asked in order to measure whether an RNMT+ decoder can distinguish
%two channels of information, one stateful features the other rich vertical features,
%we designed the next hybrid by horizontally mixing the components,
%with the hope of benefiting from diverse encoding properties of available encoders,
%at the same level.
As illustrated in Fig.~\ref{fig:mcol}, a multi-column encoder merges
the outputs of several independent encoders into a single
combined representation.
%before it is fed to the decoder.
Unlike a cascaded encoder, the multi-column encoder enables us to investigate
whether an RNMT+ decoder can distinguish information received
from two different channels and benefit from its combination.
A crucial operation in a multi-column encoder is therefore how
different sources of information are merged into
a unified representation. Our best multi-column encoder performs a simple
concatenation of individual column outputs.

%Different from Cascaded Encoder, where stateful features are enhanced/enriched using
%further vertical feature extraction
%(in the essense similar to Deep-Output of
%\cite{DBLP:journals/corr/PascanuGCB13,D17-1300}),
%multi-column encoder
%experiments aim to measure, can RNMT+ decoder benefit from individual
%streams of bare information streams, rich features from one stream,
%stateful features from the other.

The model details and hyperparameters
of the above two encoders are described in
Appendix~\ref{vertical_mixing} and \ref{horizontal_mixing}. As shown
in Table~\ref{table:hybrids-perf}, the multi-column encoder followed
by an RNMT+ decoder achieves better results than the
Transformer and the RNMT model on both WMT'14 benchmark tasks.
%we can see that an RNMT+ decoder can also
%distinguish and benefit from two different streams, which can be attained to the
%newly introduced multi-head attention network in RNMT+.

%We stick to the simple concatenation operation as the merger-operator
%for the simplicity, and after
%concatenation, the combined representation is projected down the decoder
%dimension with a layer-normalized affine transformation.
%Although in this paper we only use two columns, there is no practical
%restrictions on the number of columns. It is possible to extend number of
%columns to its extreme, learning to combine
%multiple encoder representations, hopefully capturing different factors of
%variations in the input sequence. We conclude our experiments for multi-column
%encoders by demonstrating its potential, by making use of the simplest
%merger-operator, concatenation and without fine-tuning the model or
%hyperparameters.
\begin{table}[!htbp]
\centering
\begin{tabular}{ c|c|c}
 \hline
 \hline
Model & En$\rightarrow$Fr BLEU & En$\rightarrow$De  BLEU\\ \hline
Trans. Big    & 40.73 $\pm$ 0.19 & 27.94 $\pm$ 0.18   \\
RNMT+     & 41.00 $\pm$ 0.05  & 28.49  $\pm$ 0.05  \\
Cascaded & \textbf{41.67 $\pm$ 0.11} & 28.62 $\pm$ 0.06\\
 MultiCol & 41.66 $\pm$ 0.11 & \textbf{28.84 $\pm$ 0.06}\\
 \hline
\end{tabular}
\caption{Results for hybrids with cascaded encoder and multi-column encoder.}
\label{table:hybrids-perf}
\end{table}

\vspace{-5px}
\section{Conclusion}
\label{sec:conclusion}
\vspace{-5px}
%[TODO(orhanf): talk about RNN modeling the state space, good for sequence
%modeling especially for language]

%[TODO(orhanf): talk about xformer modeling global context of a sequence, good
%at spiking features, again lacking the state information].

%[TODO(orhanf): talk about convS2S normalization, gradient scaling]

%[TODO(orhanf): talk about convS2S. modeling local context and correlations,
%lacking the state information].

In this work we explored the efficacy of several architectural and training
techniques proposed in recent studies on seq2seq
models for NMT. We demonstrated that many of these techniques
are broadly applicable to multiple model architectures.
Applying these new techniques to RNMT models yields RNMT+, an enhanced RNMT
model that
%When equipped with these new techniques, dethroned RNMT models
%improve over other architectures, and this enhanced family, RNMT+,
significantly outperforms the three fundamental architectures
on WMT'14 En$\rightarrow$Fr and En$\rightarrow$De tasks.
%We apply some of
%these techniques to RNN  based MT models and propose a new
%architecture, RNMT+, which outperforms other architectures on WMT'14
%En $\rightarrow$ Fr and En $\rightarrow$ De.
%With the lessons learned,
We further presented several hybrid models developed by combining
encoders and decoders from the Transformer and RNMT+ models, and empirically
demonstrated the superiority of the Transformer encoder and the RNMT+
decoder in comparison with their counterparts. We then enhanced the encoder
architecture by horizontally and vertically mixing components borrowed from
these architectures, leading to hybrid architectures that obtain further improvements
over RNMT+.

% We believe that our work would motivate NMT researchers to further
We hope that our work will motivate NMT researchers to further
investigate
generally applicable training and optimization techniques,
and that our exploration of hybrid architectures will open
paths for new architecture search efforts for NMT.

% Although we focused on the modeling and training techniques of the most recent
% models and conducted quality comparisons only on a single-language-pair
Our focus on a standard single-language-pair translation task leaves
important open questions to be answered:
How do our new architectures compare in multilingual settings, i.e., modeling
an \textit{interlingua}? Which architecture is more efficient and
powerful in processing finer grained inputs and outputs, e.g., characters or
bytes? How transferable are the representations learned by the different
architectures to other tasks? And what are the characteristic errors that
each architecture makes, e.g., linguistic plausibility?

\ifblindreview
\else
\section*{Acknowledgments}

We would like to thank the entire Google Brain Team and Google Translate Team for their foundational contributions to this project. We would also like to thank Noam Shazeer, Ashish Vaswani, Jakob Uszkoreit, Lukasz Kaiser, and the entire Tensor2Tensor development team for their useful inputs and discussions.
\fi

%\bibliography{acl2018}
\bibliography{acl2018}
\bibliographystyle{acl_natbib}

\vfill

\pagebreak

\appendix

\section{Supplemental Material}
\label{sec:supplemental}

\subsection{ConvS2S}

For the WMT'14 En$\rightarrow$De task, both the encoder and decoder have 15
layers, with 512 hidden units in the first ten layers, 768 units in the
subsequent three layers and 2048 units in the final two layers.
The first 13 layers use kernel width 3 and the final two layers use kernel
width 1. For the WMT'14 En$\rightarrow$Fr task, both the encoder and decoder
have 14 layers, with 512 hidden units in the first five layers, 768 units
in the subsequent four layers, 1024 units in the next three layers, 2048 units
and 4096 units in the final two layers. The first 12 layers use kernel width 3
and the final two layers use kernel width 1.
We train the ConvS2S models with synchronous training using 32 GPUs.

\subsection{Transformer}

Both the encoder and the decoder have 6 Transformer layers. Transformer base
model has model dimension 512, hidden dimension 2048 and 8 attention heads.
The Transformer Big model uses model dimension 1024, hidden dimension
8192 and 16
attention heads. We group the dropout in Transformer models into four types:
\textit{input dropout} - dropout applied to the sum of token embeddings and
position encodings, \textit{residual dropout} - dropout applied to the output
of each sublayer before added to the sublayer input, \textit{relu dropout} -
dropout applied to the inner layer output after ReLU activation in each
feed-forward sub-layer, \textit{attention dropout} - dropout applied to
attention weight in each attention sub-layer. All Transformer models
use the following learning rate schedule:
\begin{equation}
  lr = \frac{r_0}{\sqrt{d_{model}}} \cdot\min\left( \frac{t+1}{p \sqrt{p}}, \frac{1}{\sqrt{(t+1)}}\right)
\label{eq:trans_lr}
\end{equation}
where $t$ is the current step, $p$ is the number of warmup steps, $d_{model}$
is the model dimension and $r_0$ is a constant to adjust the magnitude of the
learning rate.

On WMT'14 En$\rightarrow$De, the Transformer Base model employs all
four types of dropout with $dropout\_probs=0.1$. We use
$r_0=2.0$ and $p=8000$ in the learning rate schedule. For
the Transformer Big model, only residual dropout and input dropout are
applied, both with $dropout\_probs=0.3$. $r_0=3.0$ and $p=40000$
are used in the learning rate schedule.

On WMT'14 En$\rightarrow$Fr, the Base model applies only residual dropout and
input dropout, each with $dropout\_probs=0.1$. The learning rate
schedule uses $r_0=1.0$ and $p=4000$. For the big model, we apply all four
types of dropout, each with $dropout\_probs=0.1$. The learning rate schedule
uses $r_0=3.0$ and $p=40000$.

We train both Transformer base model and big model with synchronous training using 16 GPUs.

\subsection{RNMT+}

\begin{figure}[t!]
 \centering
 \includegraphics[width=0.5\textwidth]{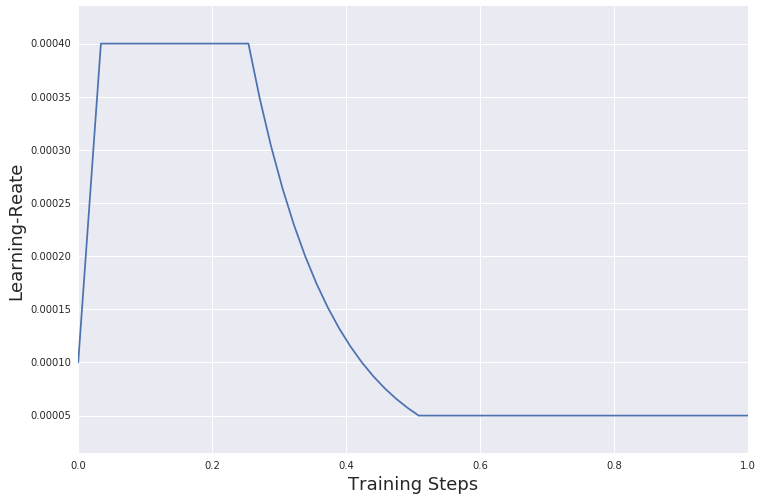}
 \caption{RNMT+ learning-rate schedule.}
 \label{fig:rnmt_lr}
 \end{figure}

RNMT+ has 1024 LSTM nodes in all encoder and decoder layers. The input
embedding dimension is 1024 as well. The encoder final projection layer
projects the last bidirectional layer output from dimension 2048 to 1024. We
use 4 attention heads in the multi-head additive attention. Label smoothing is
applied with an $uncertainty=0.1$. Figure~\ref{fig:rnmt_lr} illustrates our
learning rate schedule defined in Eq.~\ref{eq:rnmt_lr}.

On WMT'14 En$\rightarrow$De, we use $p=500$, $s=600000$, $e=1200000$ for
the learning rate schedule and apply all dropout types with
$dropout\_probs=0.3$.  We apply L2 regularization to the weights with
$\lambda=10^{-5}$. On WMT'14 En$\rightarrow$Fr, we use $p=500$,
$s=1200000$, $e=3600000$, $dropout\_probs=0.2$. No weight decay is
applied.

RNMT+ models are trained with synchronous training using 32 GPUs.

\subsection{Encoder-Decoder Hybrids}

For both encoder-decoder hybrids, i.e., Transformer Big encoder
with RNMT+ decoder and RNMT+ encoder with
Transformer Big decoder, we use the exactly same model hyperparameters
as in the Transformer Big and RNMT+ models described in above sections.

We use Transformer learning rate schedule (Eq. \ref{eq:trans_lr}) for both hybrids. For the WMT'14 En$\rightarrow$Fr task, we use $r_0=4.0$ and $p=50000$ for the hybrid with Transformer encoder and RNMT+ decoder, and use $r_0=3.0$ and $p=40000$ for the hybrid with RNMT+ encoder and Transformer decoder. Both hybrid models are trained with synchronous training using 32 GPUs.

\subsection{Cascaded Encoder Hybrid}
\label{vertical_mixing}

In this hybrid we stack a transformer encoder on top of the RNMT+
encoder. In our experiments we used a pre-trained RNMT+ encoder,
including the projection layer, exactly as described in section
\ref{sec:v2}. The outputs of the RNMT+ encoder are layer normalized
and fed into a transformer encoder.  This structure is illustrated in
Figure~\ref{fig:stacked}. The transformer encoder is identical to the
one described in subsection~\ref{subsec:transf} except for the
different number of layers. Our best setup uses 4 Transformer layers
stacked on top of a pre-trained RNMT+ encoder with 6 layers. To speed
up convergence, we froze gradient updates in the pre-trained RNMT+
encoder. This enables us to increase the encoder capacity
significantly, while avoiding optimization issues encountered in
non-frozen variants of the hybrid. As an additional benefit, this
enables us to train the model on P100s without the need for model
parallelism.

Note that this specific layout allows us to drop hand-crafted sinusoidal
positional embeddings (since position information is already captured by
the underlying RNNs).

We use the Transformer learning rate schedule (Eq. \ref{eq:trans_lr})
for this hybrid with $r_0=2.0$ and $p=16000$ and train with
synchronous training using 32 GPUs. We apply the same dropouts
used for the transformer model to the transformer layers in the
encoder, and apply L2 weight decay with $\lambda = 10^{-5}$ to the decoder layers.

\subsection{Multi-Column Encoder Hybrid}
\label{horizontal_mixing}

We use a simple concatenation as the merger-operator without
fine-tuning any other model hyperparameters. After
concatenation, the combined representation is projected down to the
decoder dimension with a layer-normalized affine transformation.
Although in this paper we only use two columns, there is no practical
restriction on the total number of columns that this hybrid can
combine. By combining multiple encoder representations, the network
may capture different factors of variations in the input sequence.
%% It is possible to extend the number of columns to its extreme, learning
%% to combine multiple encoder representations, hopefully capturing
%% different factors of variations in the input sequence. We conclude our
%% experiments for multi-column encoders by demonstrating its potential,
%% by making use of the simplest merger-operator, concatenation and
%% without fine-tuning the model or hyperparameters.

Similar to the Cascaded-RNMT+ hybrid, we use pre-trained encoders that
are borrowed from an RNMT+ model (we used a pretrained RNMT+ encoder as
the first column) and an Encoder-Decoder hybrid model with Transformer
encoder and RNMT+ decoder (we used the pretrained Transformer
encoder). Multi-column encoder with RNMT+ decoder is trained using 16
GPUs in a synchronous training setup. We stick to the simple
concatenation operation as the merger-operator, and after
concatenation, the combined representation is projected down the
decoder dimension with a simple layer-normalized affine
transformation. One additional note that we observed for the sake of
stability and trainability, each column output should be first mapped
to a space where the representation ranges are compatible, e.g., RNMT+
encoder output has not limitation on its range, but a Transformer
Encoder output range is constrained by the final layer normalization
applied to the entire Transformer encoder body. Therefore, we also
applied layer normalization to the RNMT+ encoder outputs to match the
ranges of individual encoders.

On WMT'14 En$\rightarrow$De, we use $p=50$, $s=300000$, $e=900000$ for
the learning rate schedule and apply all dropout types with
$dropout\_probs=0.3$.  We apply L2 regularization to the weights with
$\lambda=10^{-5}$. On WMT'14 En$\rightarrow$Fr, we use Transformer
learning rate schedule (Eq. \ref{eq:trans_lr}) $r_0=1.0$ and $p=10000$.
No weight decay or dropout is applied.

% \section{Multiple Appendices}

\end{document}